\documentclass[twoside,11pt]{article}

\usepackage[abbrvbib, preprint]{jmlr2e}
\hypersetup{colorlinks, linkcolor=red, citecolor = blue}
\usepackage{listings}
\usepackage{color}

\definecolor{dkgreen}{rgb}{0,0.6,0}
\definecolor{gray}{rgb}{0.5,0.5,0.5}
\definecolor{mauve}{rgb}{0.58,0,0.82}

\ShortHeadings{}{Musgrave et al.}
\firstpageno{1}

\begin{document}

\title{PyTorch Metric Learning}

\author{\name Kevin Musgrave \\
\addr Cornell Tech \\
\AND
\name Serge Belongie \\
\addr Cornell Tech \\
\AND
\name Ser-Nam Lim \\
\addr Facebook AI
}

\editor{}

\maketitle

\begin{abstract}
Deep metric learning algorithms have a wide variety of applications, but implementing these algorithms can be tedious and time consuming. PyTorch Metric Learning is an open source library that aims to remove this barrier for both researchers and practitioners. The modular and flexible design allows users to easily try out different combinations of algorithms in their existing code. It also comes with complete train/test workflows, for users who want results fast. Code and documentation is available at \href{https://www.github.com/KevinMusgrave/pytorch-metric-learning}{github.com/KevinMusgrave/pytorch-metric-learning}.
\end{abstract}


\section{Design}
\begin{figure}[h]
\centering
\includegraphics[trim={0.5cm 3cm 0.5cm 1cm},clip,width=1\textwidth]{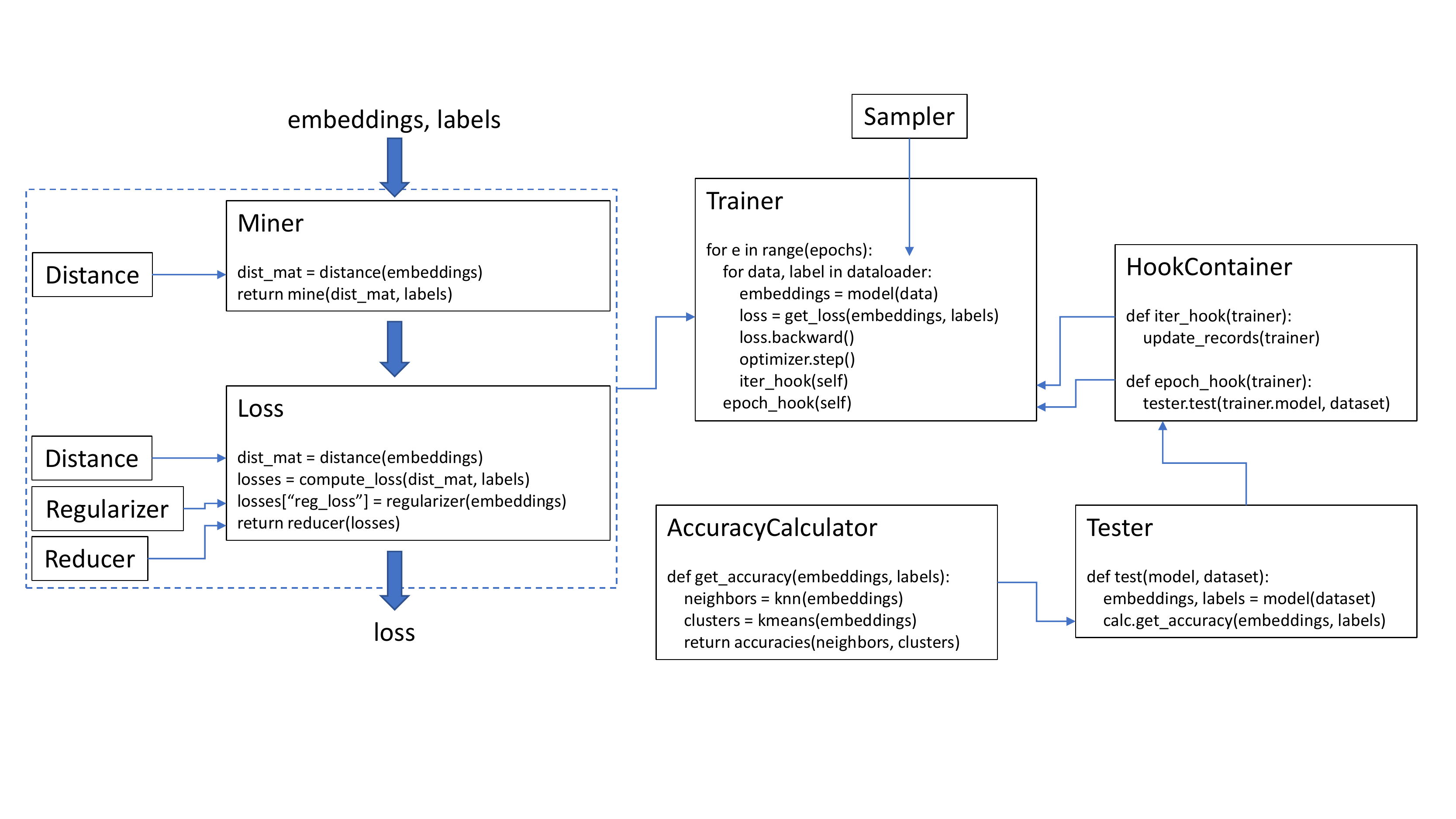}
\caption{High level view of the library's main modules, with simplified pseudo code.}
\label{overview_diagram}
\end{figure}

Figure \ref{overview_diagram} gives a high-level view of how the main modules relate to each other. Note that each module can be used independently within an existing codebase, or combined together for a complete train/test workflow. The following sections cover each module in detail.

\subsection{Losses}

\begin{figure}[h]
\centering
\includegraphics[trim={0.5cm 4cm 0.5cm 2cm},clip,width=1\textwidth]{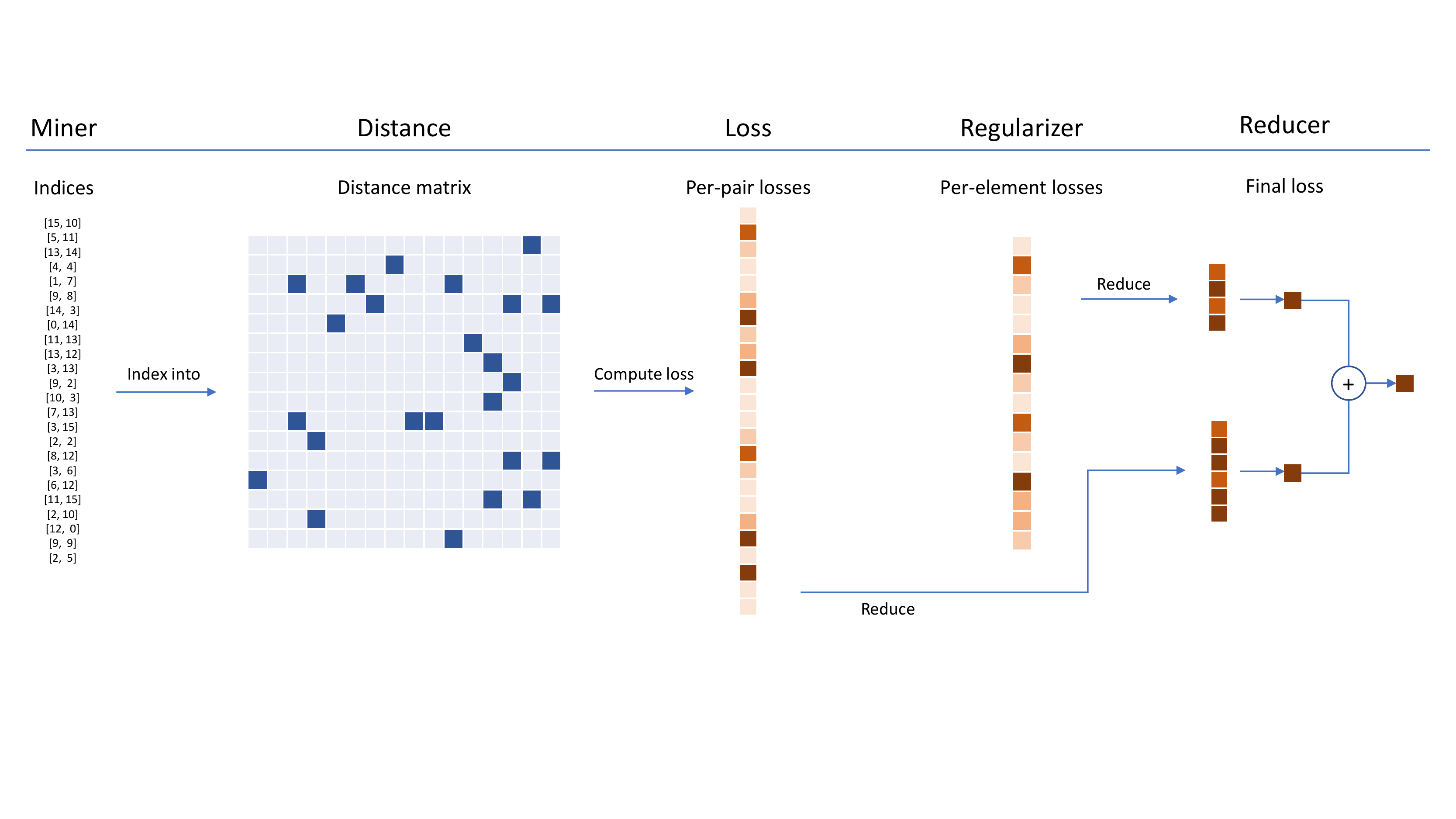}
\caption{The components of a loss function. In this illustration, a \texttt{miner} finds the indices of hard pairs in the current batch. These are used to index into the distance matrix, computed by the \texttt{distance} object. For this example, the loss function is pair-based, so it computes a loss per pair. In addition, a \texttt{regularizer} has been supplied, so a regularization loss is computed for each embedding in the batch. The per-pair and per-element losses are passed to the \texttt{reducer}, which (in this example) only keeps losses with a high value. The averages are computed for the high-valued pair and element losses, and are then added together to obtain the final loss.}
\label{loss_modularity}
\end{figure}

Loss functions work similarly to many regular PyTorch loss functions, in that they operate on a two-dimensional tensor and its corresponding labels:

\begin{lstlisting}
from pytorch_metric_learning.losses import NTXentLoss
loss_func = NTXentLoss()
### training loop ###
for data, labels in dataloader:
    embeddings = model(data)
    loss = loss_func(embeddings, labels)
    loss.backward()
\end{lstlisting}

\noindent But as shown in Figure \ref{loss_modularity}, loss functions can be augmented through the use of \texttt{miners}, \texttt{distances}, \texttt{regularizers}, and \texttt{reducers}. First consider \texttt{distances}: all losses operate on a distance matrix, whether it is the distances between each pair of embeddings in a batch, or between embeddings and learned weights. So internally, the loss function uses a \texttt{distance} object to compute a pairwise distance matrix, and then uses elements of this matrix to compute the loss.

\subsection{Distances}
As an example of how \texttt{distance} objects work, consider the \texttt{TripletMarginLoss} with its default distance metric:
\begin{lstlisting}
from pytorch_metric_learning.losses import TripletMarginLoss
loss_func = TripletMarginLoss()
\end{lstlisting}

\noindent In this form, the loss computes the following for every triplet in the batch:

\begin{equation}
L_{triplet} = [d_{ap} - d_{an} + margin]_{+}
\end{equation}

\noindent where $d$ is Euclidean distance. This distance metric can be replaced by passing in a different \texttt{distance} object:

\begin{lstlisting}
from pytorch_metric_learning.losses import TripletMarginLoss
from pytorch_metric_learning.distances import SNRDistance
loss_func = TripletMarginLoss(distance = SNRDistance())
\end{lstlisting}

\noindent Now $d$ represents the signal to noise ratio. The same loss function can also be used with inverted distance metrics, such as cosine similarity:
\begin{lstlisting}
from pytorch_metric_learning.losses import TripletMarginLoss
from pytorch_metric_learning.distances import CosineSimilarity
loss_func = TripletMarginLoss(distance = CosineSimilarity())
\end{lstlisting}

\noindent Even though \texttt{CosineSimilarity} is an inverted metric (large values indicate higher similarity), the loss function still works because it internally makes the necessary adjustments for the calculation to make sense. Specifically, the \texttt{TripletMarginLoss} swaps the anchor-positive and anchor-negative terms:

$$L_{triplet} = [s_{an} - s_{ap} + margin]_{+}$$

\noindent where $s$ is cosine similarity.

All losses, miners, and regularizers accept a \texttt{distance} argument. This makes it very easy to try out different combinations, like the \texttt{MultiSimilarityMiner} using \texttt{SNRDistance}, or the \texttt{NTXentLoss} using \texttt{LpDistance(p=1)} and so on. Note that some losses/miners/regularizers have restrictions on the type of distances they can accept. For example, some classification losses only allow \texttt{CosineSimilarity} or \texttt{DotProductSimilarity} as their distance measure between embeddings and weights. These details are available in the documentation.

\subsection{Reducers}
Losses are typically computed per element, pair, or triplet, and are then reduced to a single value by some operation, such as averaging. Many PyTorch loss functions accept a \texttt{reduction} parameter, which is usually either \texttt{"mean"}, \texttt{"sum"}, or \texttt{"none"}. In PyTorch Metric Learning, the \texttt{reducer} parameter serves a similar purpose, but with increased modularity and functionality. Specifically, a \texttt{reducer} object operates on a dictionary which describes the losses, and then returns the reduced value. For maximum flexibility, a \texttt{reducer} can be written to operate differently for per-element, per-pair, and per-triplet losses. Here is an example of how to pass a \texttt{reducer} to a loss function:

\begin{lstlisting}
from pytorch_metric_learning.losses import MultiSimilarityLoss
from pytorch_metric_learning.reducers import ThresholdReducer
loss_func = MultiSimilarityLoss(reducer = ThresholdReducer(low = 10, high = 30))
\end{lstlisting}

\noindent The \texttt{ThresholdReducer} will discard all losses that fall below \texttt{low} and above \texttt{high}, and then return the average of the remaining losses.

\subsection{Regularizers}
It is common to add embedding or weight regularization terms to the core metric learning loss. This is straightforward to do, because every loss function has an optional \\ \texttt{embedding\_regularizer} parameter:
\begin{lstlisting}
from pytorch_metric_learning.losses import ContrastiveLoss
from pytorch_metric_learning.regularizers import LpRegularizer
loss_func = ContrastiveLoss(embedding_regularizer = LpRegularizer())
\end{lstlisting}

\noindent In addition, classification losses have an optional \texttt{weight\_regularizer} parameter:
\begin{lstlisting}
from pytorch_metric_learning.losses import ArcFaceLoss
from pytorch_metric_learning.regularizers import RegularFaceRegularizer
loss_func = ArcFaceLoss(weight_regularizer = RegularFaceRegularizer())
\end{lstlisting}

\noindent The corresponding loss multipliers are specified by \texttt{embedding\_reg\_weight} and \\ \texttt{weight\_reg\_weight}.

\subsection{Miners}
An important concept in metric learning is mining, which is the process of finding the best samples to train on. Miners come in two flavors: online miners, which find the best tuples within an already sampled batch, and offline miners, which determine the best way to create batches. In this library, online miners are part of the \texttt{miners} module, while offline miners are planned to be implemented in the \texttt{samplers} module. It is easy to use an online miner in conjunction with a loss function:

\begin{lstlisting}
from pytorch_metric_learning.losses import CircleLoss
from pytorch_metric_learning.miners import MultiSimilarityMiner
loss_func = CircleLoss()
mining_func = MultiSimilarityMiner()
### training loop ###
for data, labels in dataloader:
    embeddings = model(data)
    hard_tuples = mining_func(embeddings, labels)
    loss = loss_func(embeddings, labels, hard_tuples)
    loss.backward()
\end{lstlisting}

\noindent In the above snippet, \texttt{MultiSimilarityMiner} finds the hard pairs within each batch, and passes the indices of those hard pairs to the loss function. The loss will then be computed using only those pairs. But what happens if the loss function operates on triplets and not pairs? This will still work, because the library converts tuples if necessary. Specifically:

\begin{itemize}
\item {If pairs are passed into a triplet loss, then triplets will be formed by combining each positive pair and negative pair that share the same anchor.}
\item {If triplets are passed into a pair loss, then pairs will be formed by splitting each triplet into two pairs}
\item {If pairs or triplets are passed into a classification loss, then each embedding's loss will be weighted by how frequently the embedding occurs in the pairs or triplets.}
\end{itemize}

\subsection{Samplers}
Samplers in this library are the same as PyTorch samplers, in that they are passed to dataloaders, and determine how batches are formed. Currently this module serves more as a utility than as a bank of algorithms, but in the future it will contain offline miners.

\subsection{Trainers}
Trainers exist in this library because some metric learning algorithms are more than just losses or mining functions. Some algorithms require additional networks, data augmentations, learning rate schedules etc. The goal of the \texttt{trainers} module is to provide access to these types of metric learning algorithms. In general, \texttt{trainers} make minimal assumptions, only taking care of the forward/backward pass, while leaving the choice of model, loss functions, optimizers etc. to the user. In addition, \texttt{trainers} have end-of-iteration and end-of-epoch hooks for further customizability.

\subsection{Testers}
Given a model and a dataset, a \texttt{tester} computes the embeddings, applies any specified transformations, creates visualizations of the embedding space, and determines the accuracy of the model. Accuracy calculations are performed by the aptly named \texttt{AccuracyCalculator} class. Thus, users can easily create their own accuracy metrics by passing in a custom \texttt{AccuracyCalculator} object.

\subsection{Accuracy Calculation}
The default \texttt{AccuracyCalculator} computes accuracy via its \texttt{get\_accuracy} function, and is based on k-means clustering and k-nearest neighbors (k-nn). The clustering results are used to compute Adjusted Mutual Information (AMI) and Normalized Mutual Information (NMI), while the k-nn results are used to compute Precision@1, R-Precision, and MAP@R. The output is a dictionary mapping from metric names to values. 

Writing a custom accuracy calculator is straightforward, due to the amount of boilerplate that is already provided in the parent class. Here is an example of adding a new Custom Mutual Information metric:

\begin{lstlisting}
from pytorch_metric_learning.utils import accuracy_calculator

class CustomCalculator(accuracy_calculator.AccuracyCalculator):
    
    def calculate_CMI(self, query_labels, cluster_labels, **kwargs):
        return some_complicated_function(query_labels, cluster_labels)

    def requires_clustering(self):
        return super().requires_clustering() + ["CMI"] 
\end{lstlisting}

\noindent Now \texttt{CMI} will be included in the output dictionary. This custom calculator can be used independently, or it can be passed into a \texttt{tester} object:

\begin{lstlisting}
from pytorch_metric_learning import testers
t = testers.GlobalEmbeddingSpaceTester(accuracy_calculator=CustomCalculator())
\end{lstlisting}

\subsection{Hooks}
As mentioned previously, \texttt{trainers} contain hooks that allow users to customize the end-of-iteration and end-of-epoch behavior. For users who are short of time, this library comes with the \texttt{HookContainer} class, which essentially converts \texttt{trainers} into a complete train/test workflow, with logging and model saving.

\section{Related libraries}
Other open source metric learning libraries include metric-learn (\cite{JMLR:v21:19-678}) and pyDML (\cite{JMLR:v21:19-864}). However, their focus is on classic metric learning algorithms, using numpy (\cite{walt2011numpy})  and scikit-learn (\cite{pedregosa2011scikit}). In contrast, our library focuses on deep metric learning, and uses PyTorch (\cite{paszke2019pytorch}) as its backbone.


\acks{Thank you to Ashish Shah and Austin Reiter for reviewing the code during its early stages of development, and to Chris Kruger for testing the pre-alpha version. Thanks also to open-source contributors Will Connell (wconnell), Boris Tseytlin (btseytlin), marijnl, and AlenUbuntu, for adding new features to the library. This work is supported by a Facebook AI research grant awarded to Cornell University.}


\newpage

\bibliography{references}

\end{document}